\documentclass[final]{cvpr}

\usepackage{times}
\usepackage{epsfig}
\usepackage{graphicx}
\usepackage{amsmath}
\usepackage{amssymb}
\usepackage{enumitem}
\usepackage[numbers]{natbib}
\usepackage{comment}
\usepackage{booktabs}
\usepackage{textcomp}
\usepackage{caption}
\usepackage{subcaption}
\usepackage{gensymb}
\usepackage{xcolor}
\usepackage{float}
\usepackage{dsfont}
\usepackage[pagebackref=true,breaklinks=true,colorlinks,bookmarks=false]{hyperref}

\def\cvprPaperID{6946} % *** Enter the CVPR Paper ID here
\def\confYear{CVPR 2021}
\newcommand{\TITLE}{Scan2Cap: Context-aware Dense Captioning in RGB-D Scans}
\newcommand{\METHOD}{Scan2Cap}
\newcommand{\IMPROVEMENT}{27.61}

\newcommand{\mypara}[1]{\noindent\textbf{#1}}

\newcommand{\red}[1]{{\color{red}{#1}}}

\newcommand{\blue}[1]{{\color{blue}{#1}}}
\newcommand{\orange}[1]{{\color{orange}{#1}}}

\newcommand{\best}[1]{\textbf{#1}}

\begin{document}

%%%%%%%%% TITLE
\title{\TITLE}

\author{
Dave Zhenyu Chen$^{1}$ \qquad \qquad Ali Gholami$^{2}$ \qquad \qquad Matthias Nie{\ss}ner$^{1}$ \qquad \qquad Angel X. Chang$^{2}$\\
\qquad \\
$^{1}$Technical University of Munich \qquad $^{2}$Simon Fraser University \\
% \institute{}
}

\twocolumn[{%
	\renewcommand\twocolumn[1][]{#1}%
	\maketitle
	\begin{center}
		\includegraphics[width=0.99\textwidth]{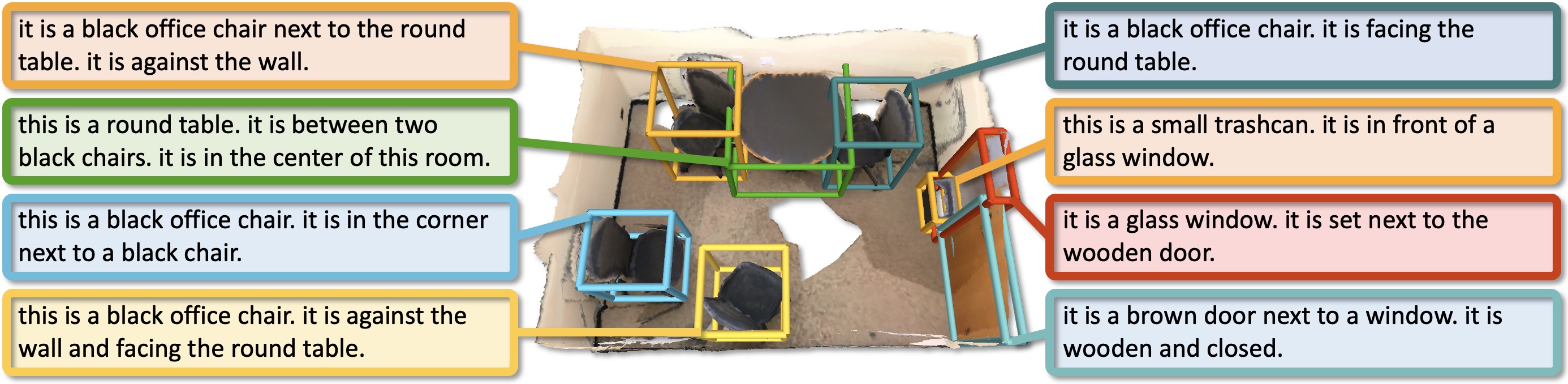}
		\captionof{figure}{
		We introduce the new task of dense captioning in RGB-D scans with a model that can densely localize objects in a 3D scene and describe them  natural language in a single forward pass.
		}
		\label{fig:teaser}
	\end{center}
}]

%%%%%%%%% ABSTRACT
% !TEX root = latex/cvpr.tex

\begin{abstract}
    We introduce the task of dense captioning in 3D scans from commodity RGB-D sensors.
    As input, we assume a point cloud of a 3D scene; the expected output is the bounding boxes along with the descriptions for the underlying objects.
    To address the 3D object detection and description problems, we propose \METHOD, an end-to-end trained method, to detect objects in the input scene and describe them in natural language.
    We use an attention mechanism that generates descriptive tokens while referring to the related components in the local context.
    To reflect object relations (i.e. relative spatial relations) in the generated captions, we use a message passing graph module to facilitate learning object relation features.
    Our method can effectively localize and describe 3D objects in scenes from the ScanRefer dataset, outperforming 2D baseline methods by a significant margin (\textbf{\IMPROVEMENT\% CiDEr@0.5IoU} improvement). 

\end{abstract}

%%%%%%%%% BODY TEXT
% !TEX root = latex/cvpr.tex

\section{Introduction}

The intersection of visual scene understanding~\citep{ren2015faster, he2017mask} and natural language processing~\citep{vaswani2017attention, devlin2018bert} is a rich and active area of research.
Specifically, there has been a lot of work on image captioning~\citep{vinyals2015show, karpathy2015deep, xu2015show, lu2017knowing, anderson2018bottom} and the related task of dense captioning~\citep{karpathy2015deep, johnson2016densecap, yang2017dense, yin2019context, kim2019dense, li2019learning}.
In dense captioning, individual objects are localized in an image and each object is described using natural language.  So far, dense captioning work has operated purely on 2D visual data, most commonly single-view images that are limited by the field of view.  Images are inherently viewpoint specific and scale agnostic, and fail to capture the physical extent of 3D objects (i.e. the actual size of the objects) and their locations in the environment.

In this work, we introduce the new task of dense captioning in 3D scenes. We aim to jointly localize and describe each object in a 3D scene. We show that leveraging the 3D information of an object such as actual object size or object location results in more accurate descriptions.

Apart from the 2D constraints in images, even seminal work on dense captioning suffers from \textit{aperture} issues~\citep{yin2019context}. Object relations are often neglected while describing scene objects, which makes the task more challenging. We address this problem with a graph-based attentive captioning architecture that jointly learns object features and object relation features on the instance level, and generates descriptive tokens. Specifically, our proposed method (referred to as \text{\METHOD}) consists of two critical components:
1) \textit{Relational Graph} facilitates learning the object features and object relation features using a message passing neural network;
2) \textit{Context-aware Attention Captioning} generates the descriptive tokens while attending to the object and object relation features.
In summary, our contribution is fourfold:
\begin{itemize}[noitemsep]
\item We introduce the 3D dense captioning task to densely detect and describe 3D objects in RGB-D scans.
\item We propose a novel message passing graph module that facilitates learning of the 3D object features and 3D object relation features.
\item We propose an end-to-end trained method that can take 3D object features and 3D object relation features into account when describing the 3D object in a single forward pass.
\item We show that our method effectively outperforms 2D-3D back-projected results of 2D captioning baselines by a significant margin (\textbf{\IMPROVEMENT\%}).

\end{itemize}
% !TEX root = latex/cvpr.tex

\begin{figure*}[!ht]
    \centering
    \includegraphics[width=\textwidth]{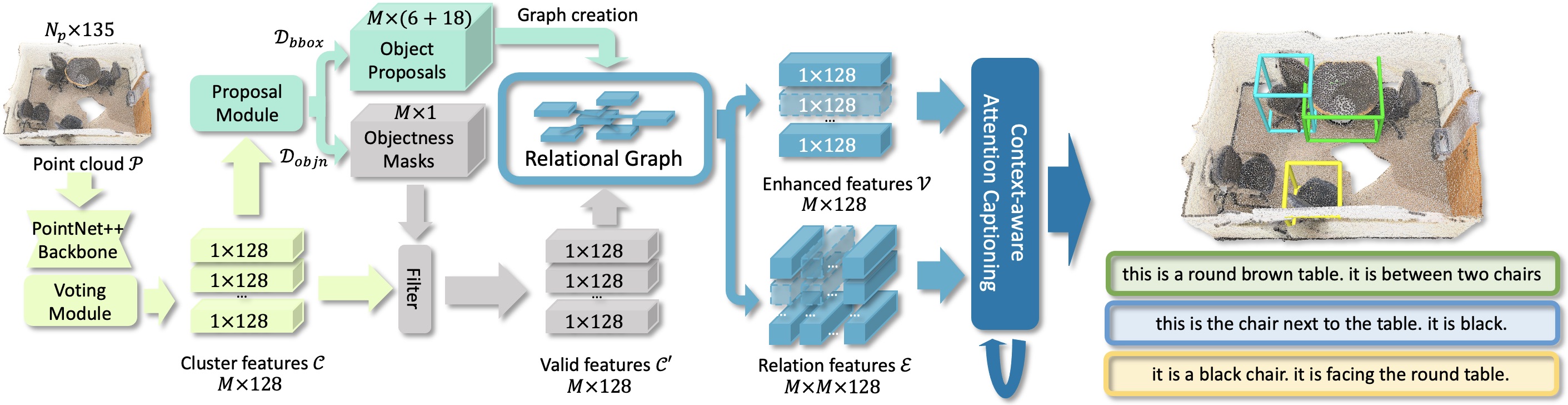}
    \caption{
    \METHOD\ takes as input a point cloud to generate the cluster features $\mathcal{C}$ for the proposal module, using a backbone following PointNet++~\citep{qi2017pointnet++} and a voting module similar to~\citet{qi2019deep}.
    The proposal module predicts the object proposals $\mathcal{D}_{\text{bbox}}$ as well as the objectness masks $\mathcal{D}_{\text{objn}}$, which are later used for filtering the cluster features as the valid features $\mathcal{C}^{'}$.
    A graph is then constructed using the object proposals and the valid cluster features.
    The relational graph module takes in the graph and outputs the enhanced object features $\mathcal{V}$ and the relation features $\mathcal{C}^{'}$.
    As the last step, the context-aware attention captioning module, inspired by~\citet{anderson2018bottom}, generates descriptive tokens for the each object proposal using the enhanced features and the relation features.
    }
    \label{fig:architecture}
\end{figure*}

\section{Related work}

\subsection{3D Object Detection} 
There are many methods for 3D object detection on 3D RGB-D datasets~\citep{song2015sun, hua2016scenenn, dai2017scannet, chang2017matterport3d}. Methods utilizing 3D volumetric grids have achieved impressive performance~\citep{hou20193d, hou2020revealnet, lahoud20193d, narita2019panopticfusion, elich20193d}. At the same time, methods operating on point clouds serve as an alternative and also achieve impressive results. For instance, \citet{qi2019deep} use a Hough voting scheme to aggregate points and generate object proposals while using a PointNet++~\citep{qi2017pointnet++} backbone. Following this work, \citet{qi2020imvotenet} recently proposed a pipeline to jointly perform voting in both point clouds and associated images. Our method builds on these works as we utilize the same backbone for processing the input geometry; however, we back-project multi-view image features to point clouds to leverage the original RGB input, since appearance is critical for accurately describing the target objects in the scene.

\subsection{Image Captioning}
Image captioning has attracted a great deal of interest~\citep{vinyals2015show, xu2015show, donahue2015long, karpathy2015deep, lu2017knowing, anderson2018bottom, jiang2018recurrent, rennie2017self}.  Attention based captioning over grid regions~\cite{xu2015show,lu2017knowing} and over detected objects~\cite{anderson2018bottom,lu2018neural} allows focusing on specific image regions while captioning.  One recent trend is the attempt to capture relationships between objects using attention and graph neural networks~\cite{gao2018image,yao2018exploring,yang2019auto} or transformers~\cite{cornia2020meshed}. We build on these ideas to propose a 3D captioning network with graphs that capture object relations in 3D.

The dense captioning task introduced by \citet{johnson2016densecap} is more closely related to our task. This task is a variant of image captioning where captions are generated for all detected objects. While achieving impressive results, this method does not consider the context outside of the salient image regions. To tackle this issue, \citet{yang2017dense} include the global image feature as context to the captioning input. \citet{kim2019dense} explicitly model the relations between detected regions in the image. Due to the limited view of a single image, prior work on 2D images could not capture the large context available in 3D environments. In contrast, we focus on decomposing the input 3D scene and capturing the appearance and spatial information of the objects in the 3D environment.

\subsection{3D Vision and Language}
While the joint field of vision and language has gained great attention in image domain, such as image captioning~\citep{vinyals2015show, xu2015show, donahue2015long, karpathy2015deep, lu2017knowing, anderson2018bottom, jiang2018recurrent, rennie2017self}, dense captioning~\citep{johnson2016densecap, yang2017dense, kim2019dense}, text-to-image generation~\citep{reed2016generative, sharma2018chatpainter, gu2018look}, visual grounding~\citep{hu2016natural, mao2016generation, yu2018mattnet}, vision and language in 3D is still not well-explored. \citet{chen2018text2shape} introduces a dataset which consists of descriptions for ShapeNet~\citep{chang2015shapenet} objects, enabling text-to-shape generation and shape captioning.
On the scene level, \citet{chen2020scanrefer} propose a dataset for localizing object in ScanNet~\citep{dai2017scannet} scenes using natural language expressions. Concurrently, \citet{achlioptasreferit3d} propose another dataset for distinguishing fine-grained objects in ScanNet scenes using natural language queries. This recent work enables research on connecting natural language to 3D environments, and inspires our work to densely localize and describe 3D objects with respect to the scene context.
% !TEX root = latex/cvpr.tex

\section{Task}

We introduce the task of dense captioning in 3D scenes. The input for this task is a point cloud of a scene, consisting of the object geometries as well as several additional point features such as RGB values and normal vectors. The expected output is the object bounding boxes for the underlying instances in the scene and their corresponding natural language descriptions.
% !TEX root = latex/cvpr.tex

\section{Method}

We propose an end-to-end architecture on the input point clouds to address the 3D dense description generation task. Our architecture consists of the following main components: 1) detection backbone; 2) relational graph; 3) context-aware attention captioning. 
As Fig.~\ref{fig:architecture} shows, our network takes a point cloud as input, and generates a set of 3D object proposals using the detection module.   A relational graph module then enhances object features using contextual cues and provides object relation features. Finally, a context-aware attention module generates descriptions from the enhanced object and relation features.

\subsection{Data Representation}\label{sec:data_representation}
As input to the detection module, we assume a point cloud $\mathcal{P}$ of one scan from ScanNet consisting of the geometry coordinates and additional point features capturing the visual appearance and the height from ground.
To obtain the extended visual point features, we follow~\citet{chen2020scanrefer} and adapt the feature projection scheme of~\citet{dai20183dmv} to back-project multi-view image features to the point cloud as additional features. The image features are extracted using a pre-trained ENet~\citep{paszke2016enet}. Following~\citet{qi2019deep}, we also append the height of the point from the ground to the new point features. As a result, we represent the final point cloud data as $\mathcal{P}=\{(p_{i}, f_i)\} \in \mathcal{R}^{N_P \times 135}$, where $p_i \in \mathcal{R}^{3}, i=1,...,N_P$ are the coordinates and $f_i \in \mathcal{R}^{132}$ are the additional features.

\subsection{Detection Backbone}
As the first step in our network, we detect all probable objects in the given point cloud with the back-projected multi-view image features discussed in~\ref{sec:data_representation}.
To construct our detection module, we adapt the PointNet++~\citep{qi2017pointnet++} backbone and the voting module in VoteNet~\citep{qi2019deep} to aggregate all object candidates to individual clusters. The output from the voting module is a set of point clusters $\mathcal{C} \in \mathcal{R}^{M \times 128}$ representing all object proposals with enriched point features, where $M$ is the upper bound of the number of proposals. Next, the proposal module takes in the point clusters to predict the objectness mask $\mathcal{D}_{\text{objn}} \in \mathcal{R}^{M \times 1}$ and the axis-aligned bounding boxes $\mathcal{D}_{\text{bbox}} \in \mathcal{R}^{M \times (6+18)}$ for all $M$ proposals, where each $\mathcal{D}_{\text{bbox}}^{i} = (c_x, c_y, c_z, r_x, r_y, r_z, l)$ consists of the box center $c$, the box lengths $r$ and a vector $l \in \mathcal{R}^{18}$ representing the semantic predictions.

\begin{figure}[!t]
    \centering
    \begin{subfigure}{\linewidth}
        \centering
        \includegraphics[width=0.99\linewidth]{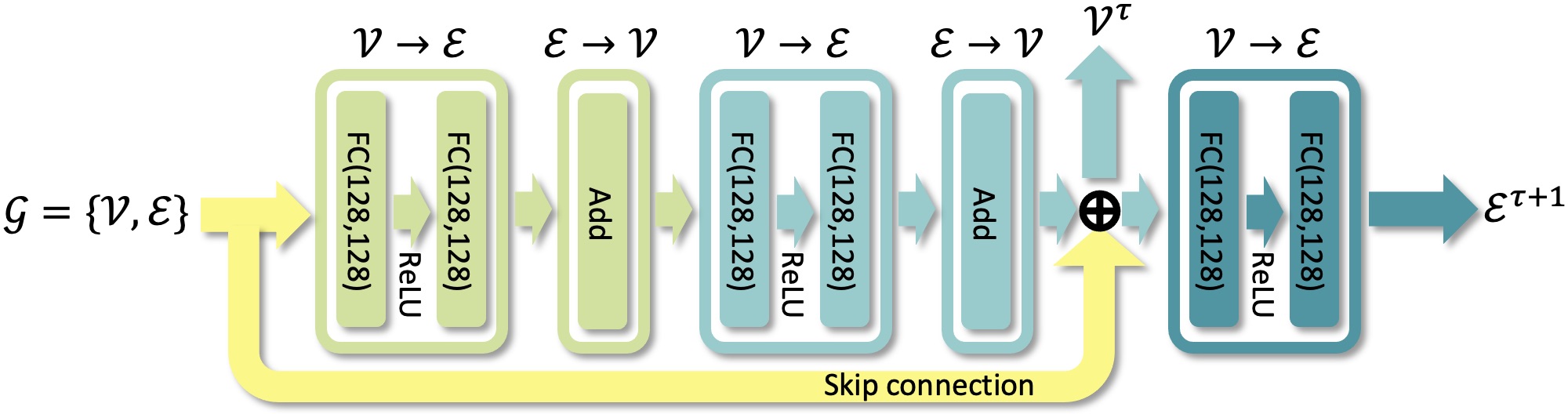}
        \caption{Relational graph module.}
        \label{fig:enhancement}
    \end{subfigure}
    \begin{subfigure}{\linewidth}
        \centering
        \includegraphics[width=0.99\linewidth]{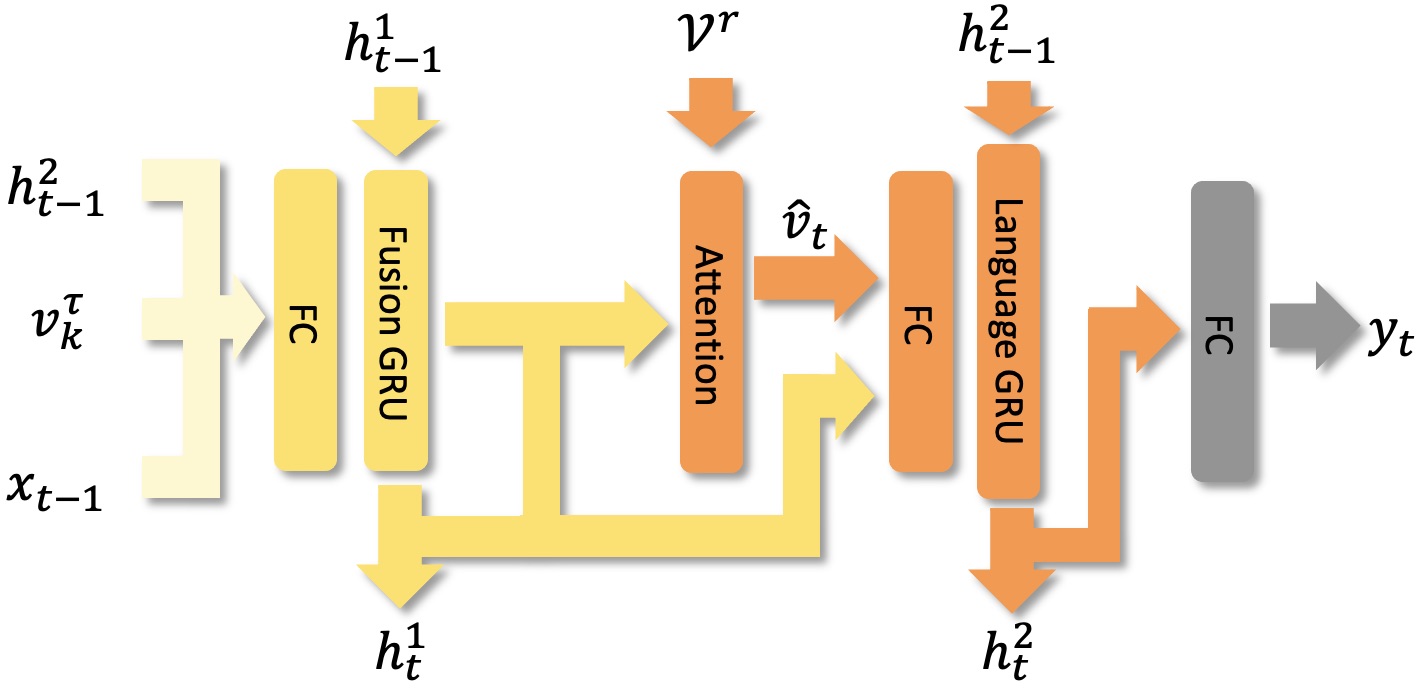}
        \caption{Context-aware attention captioning module.}
        \label{fig:attention}
    \end{subfigure}
    \caption{(a) Context enhancement module takes in the scene graph $\mathcal{G}=(\mathcal{V}, \mathcal{E})$ and produces the enhanced object features $\mathcal{V}^\tau$ and object relation features $\mathcal{E}^{\tau+1}$; (b) At time step $t$, context-aware captioning module takes in the enhanced features $v_k^\tau$ of the target object and generates the next token $y_t$ with the help of attention mechanism on the attention context features $\mathcal{V}^r$.}
    \label{fig:modules}
\end{figure}

\subsection{Relational Graph}
Describing the object in the scene often involves its appearance and spatial location with respect to nearby objects. Therefore, we propose a relational graph module equipped with a message passing network to enhance the object features and extract the object relation features.  We create a graph $\mathcal{G}=(\mathcal{V}, \mathcal{E})$ where we treat the object proposals as nodes in the graph and relationship between objects as edges.  For the edges, we consider only the nearest $K$ objects surrounding each object. We use standard neural message passing~\citep{gilmer2017neural} where the message passing at graph step $\tau$ is defined as follows:
\begin{equation}
    \mathcal{V} \to \mathcal{E}:g_{i,j}^{\tau+1}=f^{\tau}([g_i^\tau, g_j^\tau-g_i^\tau])
\end{equation}
where $g_i^\tau \in \mathcal{R}^{128}$ and $g_j^\tau \in \mathcal{R}^{128}$ are the features of nodes $i$ and $j$ at graph step $\tau$. $g_{i,j}^{\tau+1} \in \mathcal{R}^{128}$ denotes the message between nodes $i$ and $j$ at the next graph step $\tau+1$. $[\cdot, \cdot]$ concatenates two vectors. $f^\tau(\cdot)$ is a learnable non-linear function, which is in practice set as an MLP. The aggregated node features from messages after every message passing step is defined as $\mathcal{E} \to \mathcal{V}:g_{i}^{\tau+1}=\sum^K_{k=1}g_{i,k}^{\tau}$. We take the node features $\mathcal{V}^\tau$ in the last graph step $\tau$ as the output enhanced object features. We append an additional message passing layer after the last graph step and use the learned message $\mathcal{E}^{\tau+1}$ as the output object relation features. An MLP is attached to the output message passing layer to predict the angular deviations between two objects. We illustrate the relational graph module in Fig.~\ref{fig:enhancement}.

\subsection{Context-aware Attention Captioning} \label{sec:topdown_captioning}

Inspired by~\citet{anderson2018bottom}, we design a context aware attention captioning module which takes both the enhanced object features and object relation features and generates the caption one token at a time, as shown in Fig.~\ref{fig:attention}.

\mypara{Fusion GRU.}
At time-step $t$ of caption generation, we first concatenate three vectors as the fused input feature $u_{t-1}^{1}$: GRU hidden state from time-step $t-1$ denoted as $h_{t-1}^2 \in R^{512}$, enhanced object feature $v_{k}^\tau \in R^{128}$ of the $k^{th}$ object and GloVE~\citep{pennington2014glove} embedding of the token generated at $t-1$ denoted as $x_t = W_ey_{t-1} \in R^{300}$. The Fusion GRU handles the fused input feature $u_{t-1}^{1}$ and delivers the hidden state $h_t^1$ to the attention module. 

\mypara{Attention module.} 
Unlike the attention module in~\citet{anderson2018bottom} which only considers object features, we include both the enhanced object features $\mathcal{V}^\tau=\{v_i^\tau\} \in \mathcal{R}^{M \times 128}$ as well as the object relation features $e_{k, j} \in \mathcal{R}^{128}$.
We add each object relation feature $e_{k, j}$ between the object $k$ and its neighbor $j$ to the corresponding enhanced object feature $v_{j}$ of the $j^{th}$ object as the final attention context feature set $\mathcal{V}^r=\{v_{1}^r,...,v_{k}^\tau,...,v_{M}^r\}$. Intuitively, the attention module will attend to the neighbor objects and their associated relations with the current object. We define the intermediate attention distribution $\alpha_t \in \mathcal{R}^{M \times 128}$ over the context features as: 
\begin{equation}
    \alpha_t = \text{softmax}((\mathcal{V}^r W_v + \mathds{1}_{h}h_{t-1}^{1T}W_h)W_a)\mathds{1}_{a}
\end{equation}
where $W_a \in \mathcal{R}^{128 \times 1}$, $W_v \in \mathcal{R}^{128 \times 128}$, $W_h \in \mathcal{R}^{512 \times 128}$ are learnable parameters. $\mathds{1}_{h} \in \mathcal{R}^{M \times 1}$ and $\mathds{1}_{a} \in \mathcal{R}^{1 \times 128}$ are identity matrices. Finally, the attention module outputs aggregated context vector $\hat{v_t}=\sum^{M}_{i=1}\mathcal{V}^r_i \odot \alpha_{ti} $ to represent the attended object and corresponding inter-object relation.

\mypara{Language GRU.} 
We then concatenate the hidden state $h_{t-1}^1$ of the Fusion GRU in last time step and the aggregated context vector $\hat{v_t}$, and process them with a MLP as the fused feature $u_t^2$. The language GRU takes in the fused input $u_t^2$ and delivers the hidden state $h_t^2$ to the output MLP to predict token $y_{t}$ at the current time step $t$.

\subsection{Training Objective}

\mypara{Object detection loss}
We use the same detection loss $\mathcal{L}_{det}$ as introduced in~\citet{qi2019deep} for object proposals $\mathcal{D}_{\text{bbox}}$ and $\mathcal{D}_{\text{objn}}$: $\mathcal{L}_{\text{det}} = \mathcal{L}_{\text{vote-reg}} + 0.5\mathcal{L}_{\text{objn-cls}} + \mathcal{L}_{\text{box}} + 0.1\mathcal{L}_{\text{sem-cls}}$, where $\mathcal{L}_{\text{vote-reg}}$, $\mathcal{L}_{\text{objn-cls}}$, $\mathcal{L}_{\text{box}}$ and $\mathcal{L}_{\text{sem-cls}}$ represent the vote regression loss (defined in~\citet{qi2019deep}), the objectness binary classification loss, box regression loss and the semantic classification loss for the $18$ ScanNet benchmark classes, respectively. We ignore the bounding box orientations in our task and simplify $\mathcal{L}_{\text{box}}$ as $\mathcal{L}_{\text{box}} = \mathcal{L}_{\text{center-reg}} + 0.1\mathcal{L}_{\text{size-cls}} + \mathcal{L}_{\text{size-reg}}$, where $\mathcal{L}_{\text{center-reg}}$, $\mathcal{L}_{\text{size-cls}}$ and $\mathcal{L}_{\text{size-reg}}$ are used for regressing the box center, classifying the box size and regressing the box size, respectively. We refer readers to~\citet{qi2019deep} for more details.   

\mypara{Relative orientation loss}\label{sec:rel_loss}
To stabilize the learning process of the relational graph module, we apply a relative orientation loss $\mathcal{L}_{\text{ad}}$ on the message passing network as a proxy loss. 
We discretize the output angular deviations ranges from $0^{\circ}$ to $180^{\circ}$ into 6 classes, and use a cross entropy loss as our classification loss. We construct the ground truth labels using the transformation matrices of the aligned CAD models in Scan2CAD~\citep{avetisyan2019scan2cad}, and mask out objects not provided in Scan2CAD in the loss function.

\mypara{Description loss}
The main objective loss constrains the description generation. We apply a conventional cross entropy loss function $\mathcal{L}_{\text{des}}$ on the generated token probabilities, as in previous work~\citep{xu2015show, vinyals2015show, karpathy2015deep}.

\mypara{Final loss}
We combine all three loss terms in a linear manner as our final loss function: 
\begin{equation}
    \mathcal{L} = \alpha\mathcal{L}_{\text{det}} + \beta\mathcal{L}_{\text{ad}} + \gamma\mathcal{L}_{\text{des}}   
\end{equation}
where $\alpha$, $\beta$ and $\gamma$ are the weights for the individual loss terms. After fine-tuning on the validation split, we set those weights to $\alpha = 10$, $\beta = 1$, and $\gamma = 0.1$ in our experiments to ensure the loss terms are roughly of the same magnitude.

\subsection{Training and Inference}
In our experiments, we randomly select 40,000 points from ScanNet mesh vertices. During training, we set the upper bound of the number of object proposals as $M=256$. We only use the unmasked predictions corresponding to the provided objects in Scan2CAD for minimizing the relative orientation loss, as stated in~\ref{sec:rel_loss}. To optimize the description loss, we select the generated description of the object proposal with the largest IoU with the ground truth bounding box. During inference, we apply a non-maximum suppression module to suppress overlapping proposals. %There are hence way less object-description pairs generated during inference.

\subsection{Implementation Details}
We implement our architecture using PyTorch~\cite{pytorch} and train end-to-end using ADAM~\citep{kingma2014adam} with a learning rate of $1$e$-3$.
We train the model for $~90,000$ iterations until convergence.
To avoid overfitting, we set the weight decay factor to $1$e$-5$ and apply data augmentation to our training data.  Following ScanRefer~\cite{chen2020scanrefer}, the point cloud is rotated by a random angle in $[-5^{\circ},5^{\circ}]$ about all three axes and randomly translated within $0.5$ meters in all directions.
Since the ground alignment in ScanNet is imperfect, the rotation is around all axes (not just up).
We truncate descriptions longer than $30$ tokens and add SOS and EOS tokens to indicate the start and end of the description.
 
% !TEX root = latex/cvpr.tex

\begin{table*}[!ht]
    \centering
    \resizebox{\linewidth}{!}{
        \begin{tabular}{l|l|l|cccc|cccc|c}
            \toprule
            %\multicolumn{11}{c}{Description Generation in 2D} \\
            %\midrule
             & Captioning & Detection & C@0.25IoU & B-4@0.25IoU & M@0.25IoU & R@0.25IoU & C@0.5IoU & B-4@0.5IoU & M@0.5IoU & R@0.5IoU & mAP@0.5IoU \\
            \midrule
            2D-3D Proj. & 2D & Mask R-CNN & 18.29 & 10.27 & 16.67 & 33.63 & 8.31 & 2.31 & 12.54 & 25.93 & 10.50 \\
            3D-2D Proj. & 2D & VoteNet & 19.73 & 17.86 & 19.83 & 40.68 & 11.47 & 8.56 & 15.73 & 31.65 & 31.83 \\
            \midrule
            %\multicolumn{11}{c}{Description Generation in 3D} \\
            %\midrule
             %& Detection & C@0.25IoU & B-4@0.25IoU & M@0.25IoU & R@0.25IoU & C@0.5IoU & B-4@0.5IoU & M@0.5IoU & R@0.5IoU & mAP@0.5IoU \\
            %\midrule
            VoteNetRetr~\citep{qi2019deep} & 3D & VoteNet & 15.12 & 18.09 & 19.93 & 38.99 & 10.18 & 13.38 & 17.14 & 33.22 & 31.83 \\
            Ours & 3D & VoteNet & \textbf{56.82} & \textbf{34.18} & \textbf{26.29} & \textbf{55.27} & \textbf{39.08} & \textbf{23.32} & \textbf{21.97} & \textbf{44.78} & \textbf{32.21} \\
            \bottomrule
        \end{tabular}
    }
    \caption{Comparison of 3D dense captioning results obtained by Scan2Cap and other baseline methods. We average the scores of the conventional captioning metrics, e.g. CiDEr~\citep{vedantam2015cider}, with the percentage of the predicted bounding boxes whose IoU with the ground truth are greater than 0.25 and 0.5. Our method outperforms all baselines with a remarkable margin.}
    \label{tab:comp_baseline}
\end{table*}

\begin{table}[!ht]
    \centering
    \resizebox{\linewidth}{!}{
        \begin{tabular}{l|l|cccc}
            \toprule
            %\multicolumn{5}{c}{Description Generation in 2D with GT Detections} \\
            %\midrule
             & Cap & C@0.5IoU & B-4@0.5IoU & M@0.5IoU & R@0.5IoU \\
            \midrule
            OracleRetr2D & 2D & 20.51 & 20.17 & 23.76 & 50.98 \\
            Oracle2Cap2D & 2D & 58.44 & 37.05 & 28.59 & 61.35 \\
            \midrule
            %\multicolumn{5}{c}{Description Generation in 3D with GT Detections} \\
            %\midrule
            % & C@0.5IoU & B-4@0.5IoU & M@0.5IoU & R@0.5IoU \\
            %\midrule
            OracleRetr3D & 3D & 33.03 & 23.36 & 25.80 & 52.99 \\
            Oracle2Cap3D & 3D & \textbf{67.95} & \textbf{41.49} & \textbf{29.23} & \textbf{63.66} \\
            \bottomrule
        \end{tabular}
    }
    \caption{Comparison of 3D dense captioning results obtained by ours and other baseline methods with GT detections. We average the scores of the conventional captioning metrics with the percentage of the predicted bounding boxes whose IoU with the ground truth are greater than 0.5. Our method with GT bounding boxes outperforms all variants with a remarkable margin.}
    \label{tab:comp_skyline}
\end{table}

\section{Experiments}
\paragraph{Dataset.}
We use the ScanRefer~\citep{chen2020scanrefer} dataset which consists of 51,583 descriptions for 11,046 objects in 800 ScanNet~\citep{dai2017scannet} scenes. The descriptions contain information about the appearance of the objects (e.g. ``this is a black wooden chair''), and the spatial relations between the annotated object and nearby objects (e.g. ``the chair is placed at the end of the long dining table right before the TV on the wall'').

\paragraph{Train\&val splits.}
Following the official ScanRefer~\citep{chen2020scanrefer} benchmark split, we divide our data into train/val sets with 36,665 and 9,508 samples respectively, ensuring disjoint scenes for each split. Results and analysis are conducted on the val split, as the hidden test set is not officially available.

\paragraph{Metrics.}
To jointly measure the quality of the generated description and the detected bounding boxes, we evaluate the descriptions by combining standard image captioning metrics such as CiDEr~\citep{vedantam2015cider} and BLEU~\citep{papineni2002bleu}, with Intersection-over-Union (IoU) scores between predicted bounding boxes and the target bounding boxes. We define our combined metrics as $m@k\text{IoU}=\frac{1}{N}\sum^{N}_{i=0}m_{i}u_{i}$, where $u_i \in \{0,1\}$ is set to $1$ if the IoU score for the $i^{th}$ box is greater than $k$, otherwise 0. We use $m$ to represent the captioning metrics CiDEr~\citep{vedantam2015cider}, BLEU-4~\citep{papineni2002bleu}, METEOR~\citep{banerjee2005meteor} and ROUGE~\citep{lin2004rouge}, abbreviated as C, B-4, M, R, respectively. $N$ is the number of ground truth or detected object bounding boxes. We use mean average precision (mAP) thresholded by IoU as the object detection metric.

\begin{figure}[!t]
    \centering
        \centering
        \includegraphics[width=\linewidth]{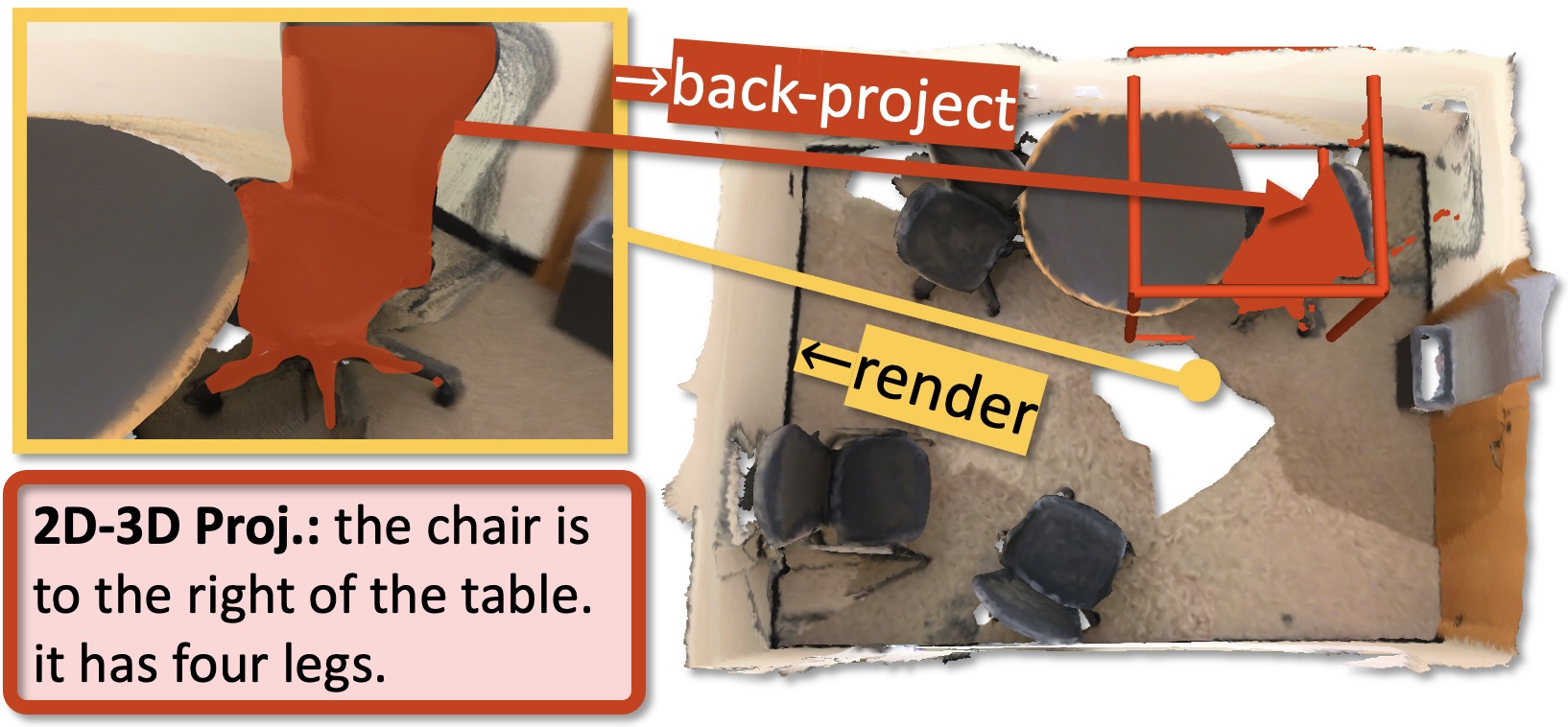}
        \caption{In 2D-3D Proj, we first generate a description for each detected object in a rendered viewpoint. Then we back-project the object mask to the 3D space to evaluate the caption with our proposed caption evaluation metric.}
        \label{fig:projection}
\end{figure}

\paragraph{Skylines with ground truth input.} To examine the upper limit of our proposed 3D dense captioning task, we use the ground truth (GT) object bounding boxes for generating object descriptions using our method and retrieval based approaches. We compare the performance of captioning in 3D with existing 2D-based captioning methods. For our 2D-based baselines, we generate descriptions for the 2D renders of the reconstructed ScanNet~\citep{dai2017scannet} scenes using the recorded viewpoints in ScanRefer~\citep{chen2020scanrefer}.

\mypara{\emph{Oracle2Cap3D}} We use ground truth 3D object bounding box features instead of detection backbone predictions to generate object descriptions. The relational graph and context-aware attention captioning module learn and generate corresponding captioning for each object. We use the same hyper-parameters with the Scan2Cap experiment.

\mypara{\emph{OracleRetr3D}} We use the ground truth 3D object bounding box features in the val split to obtain the description for the most similar object features in the train split.

\mypara{\emph{Oracle2Cap2D}} We first concatenate the global image and target object features and feed it to a caption generation method similar to~\citep{vinyals2015show}. In addition to~\citep{vinyals2015show}, we try a memory augmented meshed transformer~\citep{cornia2020meshed}. Surprisingly, the former performs better (see supplementary for details). We suspect that this performance gap is due to noisy 2D input and the size of our dataset, which does not allow for training complex methods (e.g. transformers) to their maximum potential.
The target object bounding boxes are extracted using rendered ground truth instance masks and their features are extracted using a pre-trained ResNet-101~\citep{he2016deep}. 

\mypara{\emph{OracleRetr2D}} Similar to \textit{OracleRetr3D}, use ground truth 2D object bounding box features in the val split to retrieve the description from the most similar train split object.

\paragraph{Baselines.}
We design experiments that leverage the detected object information in the input for description generation.  Additionally, we show how existing 2D-based captioning methods perform in our newly proposed task. 

\mypara{\emph{VoteNetRetr}~\citep{qi2019deep}} Similar to \textit{OracleRetr3D}, but we use the features of the 3D bounding boxes detected using a pre-trained  VoteNet~\citep{qi2019deep}.

\mypara{\emph{2D-3D Proj}} We first detect the object bounding boxes in rendered images using a pre-trained Mask R-CNN~\citep{he2017mask} with a ResNet-101~\citep{he2016deep} backbone, then feed the 2D object bounding box features to our description generation module similar to~\citet{vinyals2015show}. We evaluate the generated captions in 3D by back-projecting the 2D masks to 3D using inverse camera extrinsics (see  Fig.~\ref{fig:projection}).

\mypara{\emph{3D-2D Proj}} We first detect the object bounding boxes in scans using a pre-trained VoteNet~\citep{qi2019deep}, then project the bounding boxes to the rendered images. The 2D bounding box features are fed to our captioning module which uses the same decoding scheme as in~\citet{vinyals2015show}.

\begin{figure*}[!ht]
    \centering
    \includegraphics[width=\textwidth]{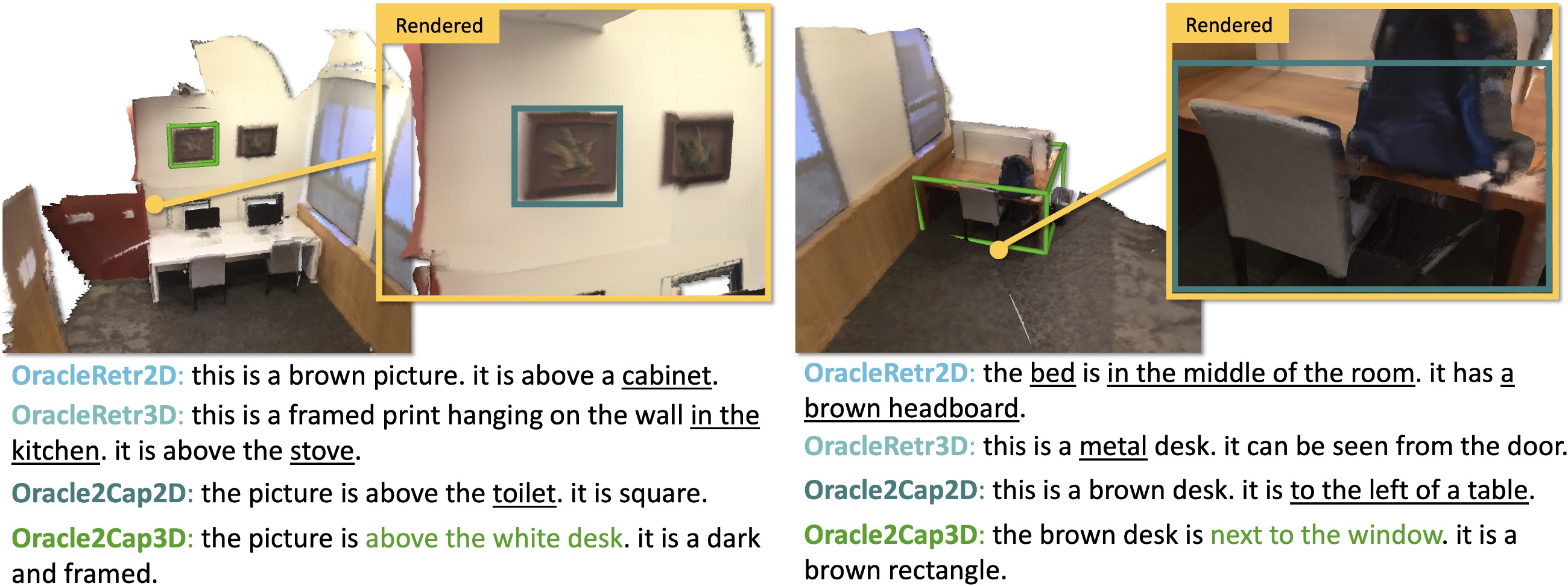}
    \caption{Qualitative results from skylines with GT input with inaccurate parts of the generated caption underscored. Captioning in 3D benefits from the richness of 3D context, while captioning with 2D information fails to capture the details of the local physical environment. Best viewed in color.}
    \label{fig:qualitative_oracle}
\end{figure*}

\begin{figure*}[!ht]
    \centering
    \includegraphics[width=\textwidth]{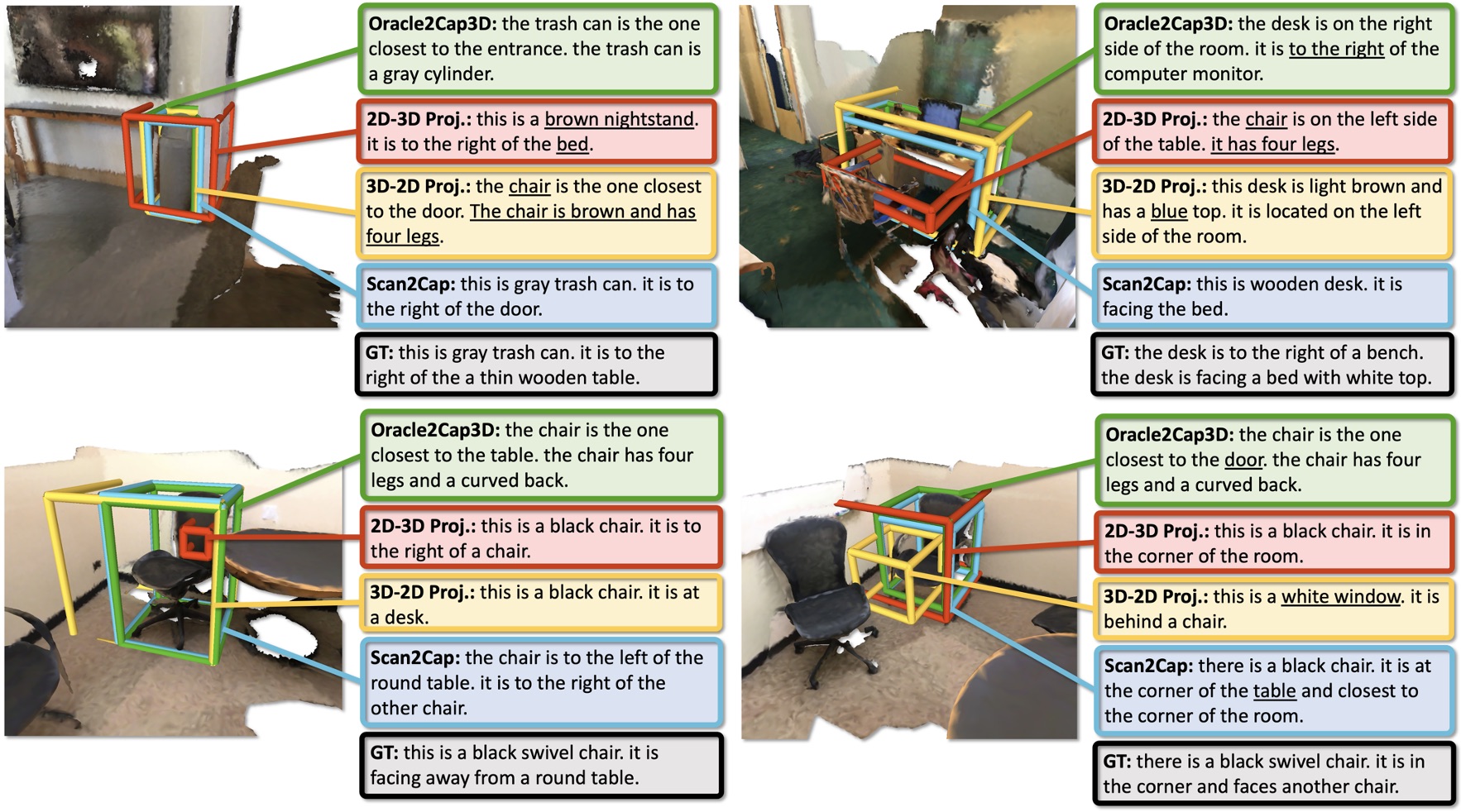}
    \caption{Qualitative results from baseline methods and \text{\METHOD} with inaccurate parts of the generated caption underscored. \blue{\METHOD} produces good bounding boxes with descriptions for the target appearance and their relational interactions with objects nearby. In contrast, the baselines suffers from poor bounding box predictions or limited view and produces less informative captions. Best viewed in color.}
    \label{fig:qualitative}
\end{figure*}

\begin{figure*}[!ht]
    \centering
    \includegraphics[width=\textwidth]{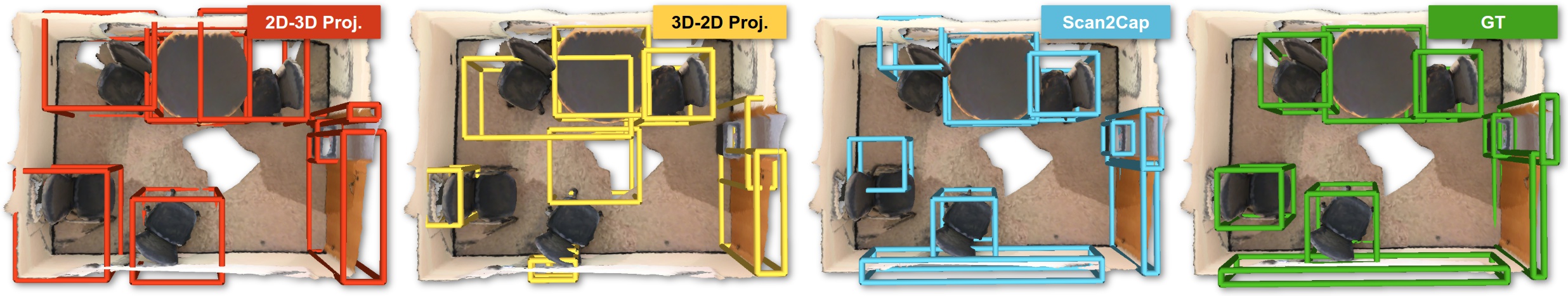}
    \caption{Comparison of object detections of baseline methods and \METHOD. \red{2D-3D Proj.} suffers from the detection performance gap between image and 3D space. \blue{\METHOD} produces better bounding boxes compared to \orange{3D-2D Proj.} due to the end-to-end fine-tuning.}
    \label{fig:detections}
\end{figure*}

\subsection{Quantitative Analysis}

We compare our method with the baseline methods on the official val split of ScanRefer~\citep{chen2020scanrefer}. As there is no direct prior work on this newly proposed task, we divide description generation into: 1) generating the object bounding boxes and descriptions in 2D input, and back-projecting the bounding boxes to 3D using camera parameters; 2) directly generating object bounding boxes with descriptions in 3D space. As shown in Tab.~\ref{tab:comp_baseline}, describing the detected objects in 3D  results in a big performance boost compared to the back-projected 2D approach (39.08\% compared to 11.47\% on C@0.5IoU). When using ground truth, descriptions generated with 3D object bounding boxes (\textit{Oracle2Cap3D}) effectively outperform their counterparts that use 2D object bounding box information (\textit{Oracle2Cap2D}), as shown in Tab.~\ref{tab:comp_skyline}. The performance gap between our method and \textit{Oracle2Cap3D} indicates that the detection backbone can be further improved as a potential future work. 

\begin{table}[t]
    \centering
    \resizebox{\linewidth}{!}{
        \begin{tabular}{l|l|ccc}
            \toprule
            & Cap & Acc (Category) & Acc (Attribute) & Acc (Relation) \\
            \midrule
            Oracle2Cap2D & 2D & 69.00 & 67.42 & 37.00 \\
            \midrule
            Oracle2Cap3D & 3D & 85.15 (\textbf{+16.15}) & 72.22 (\textbf{+4.80}) & 76.24 (\textbf{+39.24}) \\
            Ours & 3D & 84.16 (\textbf{+15.16}) & 64.21 (-3.21) & 69.00 (\textbf{+32.00}) \\
            \bottomrule
        \end{tabular}
    }
    \caption{Manual analysis of the generated captions obtained by skyline methods with GT input and ours. We measure the accuracy of three different aspects (object categories, appearance attributes and spatial relations) in the generated captions. Compared to captioning in 2D, captioning directly in 3D better capture these aspects in descriptions, especially for describing spatial relations in the local environment.}
    \label{tab:manual_analysis}
\end{table}

\begin{table*}[!ht]
    \centering
    \resizebox{\linewidth}{!}{
        \begin{tabular}{l|cccc|cccc|c}
            \toprule
            & C@0.25IoU & B-4@0.25IoU & M@0.25IoU & R@0.25IoU & C@0.5IoU & B-4@0.5IoU & M@0.5IoU & R@0.5IoU & mAP@0.5IoU\\
            \midrule
            Ours (fixed VoteNet) & 56.20 & \textbf{35.14} & 26.14 & \textbf{55.71} & 33.87 & 20.11 & 20.48 & 42.33 & 31.83\\
            Ours (end-to-end) & \textbf{56.82} & 34.18 & \textbf{26.29} & 55.27 & \textbf{39.08} & \textbf{23.32} & \textbf{21.97} & \textbf{44.78} & \textbf{32.21} \\
            \bottomrule
        \end{tabular}
    }
    \caption{Ablation study with a fixed pre-trained VoteNet~\citep{qi2019deep} and an end-to-end fine-tuned VoteNet. We compute standard captioning metrics with respect to the percentage of the predicted bounding box whose IoU with the ground truth are greater than 0.25 and 0.5. The higher the better.}
    \label{tab:comp_finetune}
    % \vspace{0.3cm}
\end{table*}

\begin{table}[!ht]
    \centering
    \resizebox{\linewidth}{!}{
        \begin{tabular}{l|cccc}
            \toprule
            & C@0.5IoU& B-4@0.5IoU & M@0.5IoU & R@0.5IoU \\
            \midrule
            VoteNet~\citep{qi2019deep}+GRU~\citep{chung2014empirical} & 34.31 & 21.42 & 20.13 & 41.33 \\
            VoteNet~\citep{qi2019deep}+CAC  & 36.15 & 21.58 & 20.65 & 41.78 \\
            VoteNet~\citep{qi2019deep}+RG+CAC & \textbf{39.08} & \textbf{23.32} & \textbf{21.97} & \textbf{44.78} \\
            \bottomrule
        \end{tabular}
    }
    \caption{Ablation study with different components in our method: VoteNet~\citep{qi2019deep} + GRU~\citep{chung2014empirical}, which is similar to ``show and tell''~\citep{vinyals2015show}; VoteNet + Context-aware Attention Captioning (CAC); VoteNet + Relational Graph (RG) + Context-aware Attention Captioning (CAC), namely \METHOD. We compute standard captioning metrics with respect to the percentage of the predicted bounding boxes whose IoU with the ground truth are greater than 0.5. The higher the better. Clearly, our method with attention mechanism and graph module is shown to be effective.}
    \label{tab:comp_component}
\end{table}

\subsection{Qualitative Analysis}
We see from Fig.~\ref{fig:qualitative_oracle} that the captions retrieved by OracleRetr2D hallucinates objects that are not there, while Oracle2Cap2D provides inaccurate captions that fails to capture correct local context.  In contrast, the captions from Oracle2Cap3D is longer and capture relationships with the surrounding objects, such as the ``above the white desk'' and ``next to the window''.  Fig.~\ref{fig:qualitative} show the qualitative results of Oracle2Cap3D, 2D-3D Proj, 3D-2D Proj and our method (\METHOD). Leveraging the end-to-end training, \text{\METHOD} is able to predict better object bounding boxes compared to the baseline methods (see Fig.~\ref{fig:qualitative} top row). Aside from the improved quality of object bounding boxes, descriptions generated by our method are richer when describing the relations between objects (see second row of Fig.~\ref{fig:qualitative}).

Provided with the ground truth object information, Oracle2Cap3D can include even more details in the descriptions. However, there are mistakes with the local surroundings (see the sample in the right column in Fig.~\ref{fig:qualitative}), indicating there is still room for improvement. In contrast, image-based 2D-3D Proj. suffers from limitations of the 2D input and fails to produce good bounding boxes with detailed descriptions. Compared to our method, 3D-2D Proj. fails to predict good bounding boxes because of the lack of a fine-tuned detection backbone, as shown in Fig.~\ref{fig:detections}.

\subsection{Analysis and Ablations}

\paragraph{Is it better to caption in 3D or 2D?} 

One question we want to study is whether it is better to caption in 3D or 2D. Therefore, we conduct a manual analysis on randomly selected 100 descriptions generated by Oracle2Cap2D, Oracle2Cap3D and our method. In this analysis, we manually check if those descriptions correctly capture three important aspects for indoor objects: object categories, appearance attributes and spatial relations in local environment. As demonstrated in Tab.~\ref{tab:manual_analysis}, directly captioning objects in 3D captures those aspects more accurately when comparing Oracle2Cap3D with Oracle2Cap2D, especially for describing the spatial relations. However, the accuracy drop on object attributes from Oracle2Cap2D to our method (-3.21\%) shows the detection backbone can still be improved.

\paragraph{Does context-aware attention captioning help?}
We compare our model with the basic description generation component (GRU) introduced in~\citet{vinyals2015show} and our model with the context-aware attention captioning (CAC) as discussed in Sec.~\ref{sec:topdown_captioning}. The model equipped with the context-aware captioning module outperforms its counterpart without attention mechanism on all metrics (see the first row vs. the second row in Tab.~\ref{tab:comp_component}).

\paragraph{Does the relational graph help?}
We evaluate the performance of our method against our model without the proposed relational graph (RG) and/or the context-aware attention captioning (CAC). As shown in Tab.~\ref{tab:comp_component}, our model equipped with context enhancement module (third row) outperforms all other ablations.

\paragraph{Does end-to-end training help?} 
We show in Tab.~\ref{tab:comp_finetune} the effectiveness of fine-tuning the pretrained VoteNet end-to-end with the description generation objective. We observe that end-to-end training of the network allows for gradient updates from our relative orientation loss and description generation loss that compensate for the detection errors. While the fine-tuned VoteNet detection backbone delivers similar detection results, its performance on describing objects outperforms its fixed ablation by a big margin on all more demanding metrics (see columns for metrics $m@0.5\text{IoU}$ in Tab.~\ref{tab:comp_finetune}).

% !TEX root = latex/cvpr.tex

\section{Conclusion}

    In this work, we introduce the new task of dense description generation in RGB-D scans.
    We propose an end-to-end trained architecture to localize the 3D objects in the input point cloud and generate the descriptions for them in natural language, which is able to address the 3D localization and describing problem at the same time.
    We apply an attention-based captioning pipeline equipped with a message passing network to generate descriptive tokens while referring to the related components in the local context.
    Our architecture can effectively localize and describe the 3D objects in the scene and it also outperforms the 2D-based dense captioning methods on the 3D dense description generation task by a big margin.
    Overall, we hope that our work will enable future research in the 3D visual language field.
\section*{Acknowledgements}

This work is funded by Google (AugmentedPerception), the ERC Starting Grant Scan2CAD (804724), and a Google Faculty Award. We would also like to thank the support of the TUM-IAS Rudolf M{\"o}{\ss}bauer and Hans Fischer Fellowships (Focus Group Visual Computing), as well as the the German Research Foundation (DFG) under the Grant \textit{Making Machine Learning on Static and Dynamic 3D Data Practical}.
Angel X. Chang is supported by the Canada CIFAR AI Chair program, and Ali Gholami is funded by an NSERC Discovery Grant.

{\small
\setlength{\bibsep}{0pt}
\bibliography{main.bib}
}

\clearpage
% \newpage

\appendix
\section*{Supplementary Material}

In the supplemental, we provide additional details on the 2D captioning experiments to explain the choice of 2D input and captioning method that we use (Sec.~\ref{sec:2dexpr}).  We also provide details about the 3d-to-2d projection (Sec.~\ref{sec:3d2dproj_details}), additional ablation studies (Sec.~\ref{sec:add_quantitative}) and qualitative examples (Sec.~\ref{sec:add_qualitative}) for our 3D experiments.

\begin{figure}[!ht]
    \centering
%    \begin{subfigure}{\textwidth}
%        \centering
        \includegraphics[width=\linewidth]{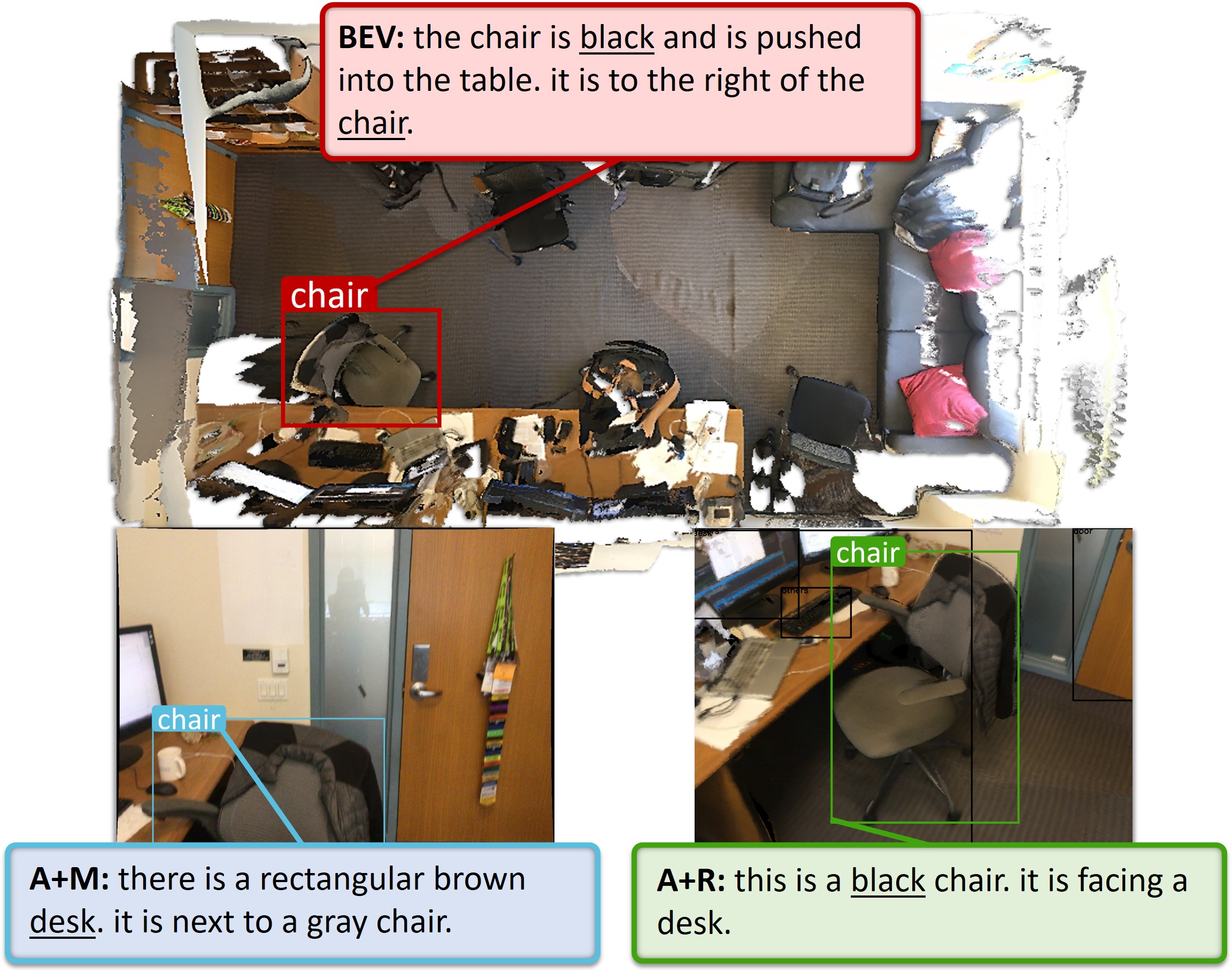}
        \caption{We compare how each input choice affects the performance of our 2D captioning experiments with oracle bounding boxes. We show the caption generated using show and tell (S\&T) for the best matching frame selected from the video recording (A+M, bottom left), rendered annotated viewpoint (A+R, bottom right), and from the bird's eye view (BEV, top). The BEV provides a good overview of large objects, but can miss smaller objects such as trashcans placed underneath desks.  The matched frame may not fully capture the object of interest or provide enough context for informative captions (see Tab.~\ref{tab:2dview_comp} for quantitative comparisons).}
        \label{fig:input_performance_comparison}
%        \end{subfigure}
    % \begin{subfigure}{\textwidth}
    %     \centering
    %     \includegraphics[width=0.99\linewidth]{figures/supplements/sorted.jpg}
    %     \caption{Examples of bad matched frames.}
    %     \label{fig:rendered_density}
    % \end{subfigure}
    % \caption{Examples of bad matches.}
    % \label{fig:modules}
\end{figure}

\section{2D experiments}
\label{sec:2dexpr}
\subsection{Experimental setup}
We conduct a series of experiments in 2D to select the input, captioning method, and visual features for our 2D baselines. We implement the models for the 2D experiments using PyTorch~\citep{pytorch} and Detectron2~\citep{wu2019detectron2}.

\begin{table}[!t]
    \centering
    \resizebox{\linewidth}{!}{
        \begin{tabular}{l|l|l|l|cccc}
            \toprule
            \multicolumn{8}{c}{Description Generation in 2D (rendered vs matched vs BEV)} \\
            \midrule
             VF & VP & DET & CAP & C & B-4 & M & R \\
            \midrule
            G  & A+R  & - & S\&T &  49.61 & 11.41 &  15.64 &  40.59  \\ 
            G  + T  & A+R & O & S\&T &  \best{59.12} & \best{12.73} & \best{16.61} & \best{41.32}  \\ 
            G  & A+M & - & S\&T & 11.50 & 1.63 & 5.64 & 13.86 \\ 
            G  + T  & A+M & O & S\&T  & 16.76 & 2.01 & 6.14 & 14.23 \\ 
            G & BEV & - & S\&T & 19.94 & 8.74 & 14.64 & 36.53  \\
            G + T & BEV & O & S\&T & 24.21 & 9.69 & 14.41 & 37.38  \\
            \midrule
            T  + C  & A+R & O & TD & \best{51.35} & \best{13.09} & \best{15.88} & \best{43.52} \\
            G  + T + C  & A+R & O & TD &  18.10 & 5.65 & 11.37 & 33.10 \\ 
            T  + C  & A+M & O & TD & 12.77 & 1.58 & 5.84 & 15.42 \\ 
            G  + T + C  & A+M & O & TD & 14.00 & 1.68 & 5.74 & 15.41 \\ 
            \bottomrule
        \end{tabular}
    }
    \caption{We compare captions for oracle bounding boxes from annotated viewpoints with rendered (A+R), matched frames (A+M), and from the birds-eye-view (BEV) on the ScanRefer~\citep{chen2020scanrefer} validation split.  We observe that the rendered frames consistently result in better captions for different features (global (G), with target object features (T), and context object features (C)) and captioning methods (show and tell (S\&T) vs top-down attention (TD)).}
    \label{tab:2dview_comp}
\end{table}

\paragraph{Choice of 2D input}

\begin{figure*}[!ht]
    \centering
%    \begin{subfigure}{\textwidth}
%        \centering
        \includegraphics[width=\linewidth]{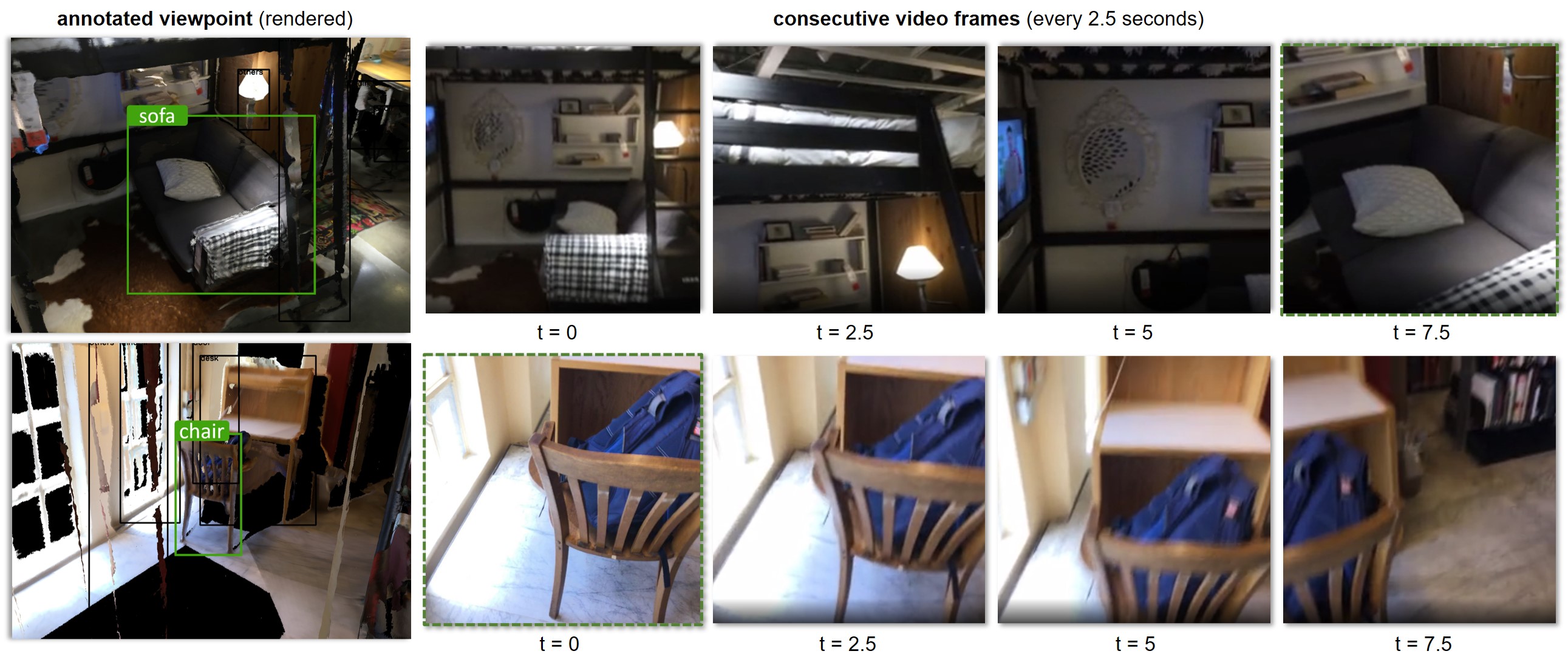}
        \caption{Examples of difficult to match viewpoints, with the rendered frame for the annotated viewpoint on the left, and sample frames from the video on the right (selected matched frame shown with dashed borders).  The bounding box for the target object is shown in green. Due to a lack of video recording coverage, it is often impossible to match the exact viewpoint camera direction and origin.  Frames from the video recording suffers from motion blur and have a view that is too close up, and missing contextual objects.}
        \label{fig:match_poor}
%        \end{subfigure}
    % \begin{subfigure}{\textwidth}
    %     \centering
    %     \includegraphics[width=0.99\linewidth]{figures/supplements/sorted.jpg}
    %     \caption{Examples of bad matched frames.}
    %     \label{fig:rendered_density}
    % \end{subfigure}
    % \caption{Examples of bad matches.}
    % \label{fig:modules}
\end{figure*}

However, we find that it is often challenging to find a good matching frame (see Fig.~\ref{fig:match_poor}), and using the rendered frames leads to better captioning performance (see Tab.~\ref{tab:2dview_comp}) despite the rendering artifacts. Fig.~\ref{fig:match_poor} shows examples of viewpoints for which it is challenging to find a good matching frame from the video frames.  We suspect that the poor performance of captioning with matched frames is due to the differences in viewpoints as well as the extremely limited field of view and motion blur found in the video frames. In addition, we also check the captioning performance from a bird-eye-view.

\paragraph{Captioning method}
For selecting a 2D captioning method, we experiment with a simple model, show and tell (S\&T~\cite{vinyals2015show}), as well as the popular bottom-up and top-down attention model (TD~\cite{anderson2018bottom}), and a recent state-of-the-art captioning method, the meshed-memory transformer (M$^2$~\cite{cornia2020meshed}). The S\&T~\citep{vinyals2015show} and TD~\citep{anderson2018bottom} models are similar to the original ones, but we replace LSTM~\citep{hochreiter1997long} with GRU~\citep{chung2014empirical} due to the small size of the ScanRefer~\citep{chen2020scanrefer} dataset. In addition to the captioning methods above, we also compare our method against the retrieval baselines (Retr).

\begin{table*}[t]
    \centering
    \resizebox{\linewidth}{!}{
        \begin{tabular}{r|cccccccccccccccccc|ccc}
            \toprule
              & bath. & bed & bkshf. & cab. & chair & cntr. & curt. & desk & door & others & pic. & fridg. & showr. & sink & sofa & tabl. & toil. & wind. & mAP & mAP50 & mAP75\\
             \midrule
%            \multicolumn{21}{c}{Object Detection}\\
%            \midrule
             DET & 12.84 & 37.66 & 20.33 & 16.09 & 32.39 & 18.63 & 16.21 & 14.47 & 14.55 & 20.98 & 24.72 & 17.30 & 18.90 & 19.73 & 29.91 & 28.71 & 58.22 & 16.09 & 23.21 & 36.01 & 24.45 \\
          \midrule
%            \multicolumn{21}{c}{Instance Segmentation}\\
%            \midrule
             SEG & 9.74 & 23.61 & 1.38 & 15.25 & 27.97 & 7.53 & 12.82 & 6.95 & 11.79 & 19.66 & 23.74 & 18.12 & 17.91 & 20.03 & 25.86 & 28.23 & 56.72 & 9.62 & 18.72 & 32.01 & 19.37\\
            \bottomrule
        \end{tabular}
    }
    \caption{2D object detection (DET) and instance segmentation (SEG) results on the ScanRefer~\citep{chen2020scanrefer} validation split. Reported values for each object category is the \textit{mAP} at IoU $=0.50:.05:0.95$ (averaged over 10 IoU thresholds). \textit{mAP} is the class averaged precision at IoU $=0.50:.05:0.95$ (averaged over 10 IoU thresholds). \textit{mAP50} is the class averaged precision at IoU $=0.50$. \textit{mAP75} is the class averaged precision at IoU $=0.75$. We use a Mask R-CNN~\citep{he2017mask} with a pre-trained ResNet-101~\cite{he2016deep} backbone and fine-tune it on the ScanRefer~\citep{chen2020scanrefer} training split.}
    \label{tab:detection_and_segmentation}
\end{table*}

\begin{table}[t]
    \centering
    \resizebox{\linewidth}{!}{
        \begin{tabular}{l|l|l|l|cccc}
            \toprule
            \multicolumn{8}{c}{Description Generation in 2D (Rendered Viewpoints)} \\
            \midrule
             VF & VP & DET & CAP & C & B-4 & M & R \\
            \midrule
            G  & A  & - & Retr &  12.07 & 4.58 &  11.50 &  29.37 \\ 
            G  & A  & - & S\&T &  49.61 & 11.41 &  15.64 &  40.59  \\ 
            \midrule
            T  & A & O & Retr & 23.00 & 7.28 & 13.44 & 33.82\\ 
            T  + C  & A & O & TD & 51.35 & \best{13.09} & 15.88 & \best{43.52}\\ 
            T  + C  & A & O & M$^2$ & 34.72 & 7.13 & 12.69 & 33.60  \\ 
            T  + C  & A & O & M$^2$ RL & 42.77 & 9.03 & 14.34 & 36.27\\ 
            G  + T  & A & O & S\&T &  \best{59.12} & 12.73 & \best{16.61} & 41.32\\ 
             G  + T + C  & A & O & TD &  18.10 & 5.65 & 11.37 & 33.10 \\ 
            \midrule
            T  + C   & A & 2DM & TD & 35.65 & \best{11.00} & 14.30 & \best{40.70}\\ 
            T  + C   & A & 2DM & M$^2$ & 31.02 & 7.19 &  12.28 &  33.22 \\
            T  + C   & A & 2DM & M$^2$ RL & 35.91 & 8.52 &  13.53 &  35.33 \\ 
            G  + T  & A & 2DM & S\&T &  \best{41.44} & 10.95 &  \best{15.08} &  39.04\\ 
            G  + T + C  & A & 2DM & TD &  14.84 & 4.95 &  10.85 &  31.52 \\ 
            \midrule
            G  & E  & - & S\&T &  28.52 & 24.03 &  18.92 &  47.76 \\ 
            T  + C  & E &  3DV & TD & 28.25 & \best{30.11} &  18.9 &  \best{52.14}  \\ 
            T  + C  & E &  3DV & M$^2$ & 11.44 & 19.67 &  14.23 &  40.42 \\ 
            T  + C  & E &  3DV & M$^2$ RL & 11.83 & 24.79 &  15.47 &  42.69 \\ 
            G  + T  & E & 3DV & S\&T &  \best{31.48} & 25.35 &  \best{19.09} &  47.06\\ 
            G  + T + C  & E &  3DV & TD & 9.66 & 9.68 &  13.14 &  38.38 \\ 
%            \midrule
%            G & BEV & - & S\&T & 19.94 & 8.74 & \best{14.64} & 36.53 & the chair is black and is sitting in front of the desk. it is to the right of the chair. \\
%            G + T & BEV & O & S\&T & \best{24.21} & \best{9.69} & 14.41 & \best{37.38} & the chair is black and is pushed into the table. it is to the right of the chair. \\
            \bottomrule
        \end{tabular}
    }
    \caption{Results of caption generation with rendered viewpoints on the ScanRefer~\citep{chen2020scanrefer} validation split. Captioning metrics are calculated by comparing the generated caption against the reference caption corresponding to the annotated viewpoint. VF is the input visual feature which can include the full image (G), context objects (C), and/or target object (T). VP is the viewpoint that can be annotated (A), estimated (E), or bird's eye viewpoint (BEV). DET is the object bounding box which can be the ground truth box (O), Mask R-CNN~\citep{he2017mask} detected in 2D (2DM) or back-projected VoteNet~\citep{qi2019deep} detection in 3D (3DV). CAP is the captioning method which can be cosine retrieval (Retr), Show and tell (S\&T)~\citep{vinyals2015show}, Top-down attention~\citep{anderson2018bottom} (TD), Meshed memory transformer~\citep{cornia2020meshed} without and with self-critical optimization respectively (M$^2$) and (M$^2$ RL).  Since S\&T with global and target object features (G+T) gives the best CiDEr score, we select it as the 2D captioning method for the main paper. }
    \label{tab:comp_baseline}
\end{table}

\begin{figure*}[!ht]
    \centering
    %begin{subfigure}{\linewidth}
    %    \centering
        %\includegraphics[width=\textwidth]{figures/supplements/qualitative_2d.jpg}
        \includegraphics[width=\textwidth]{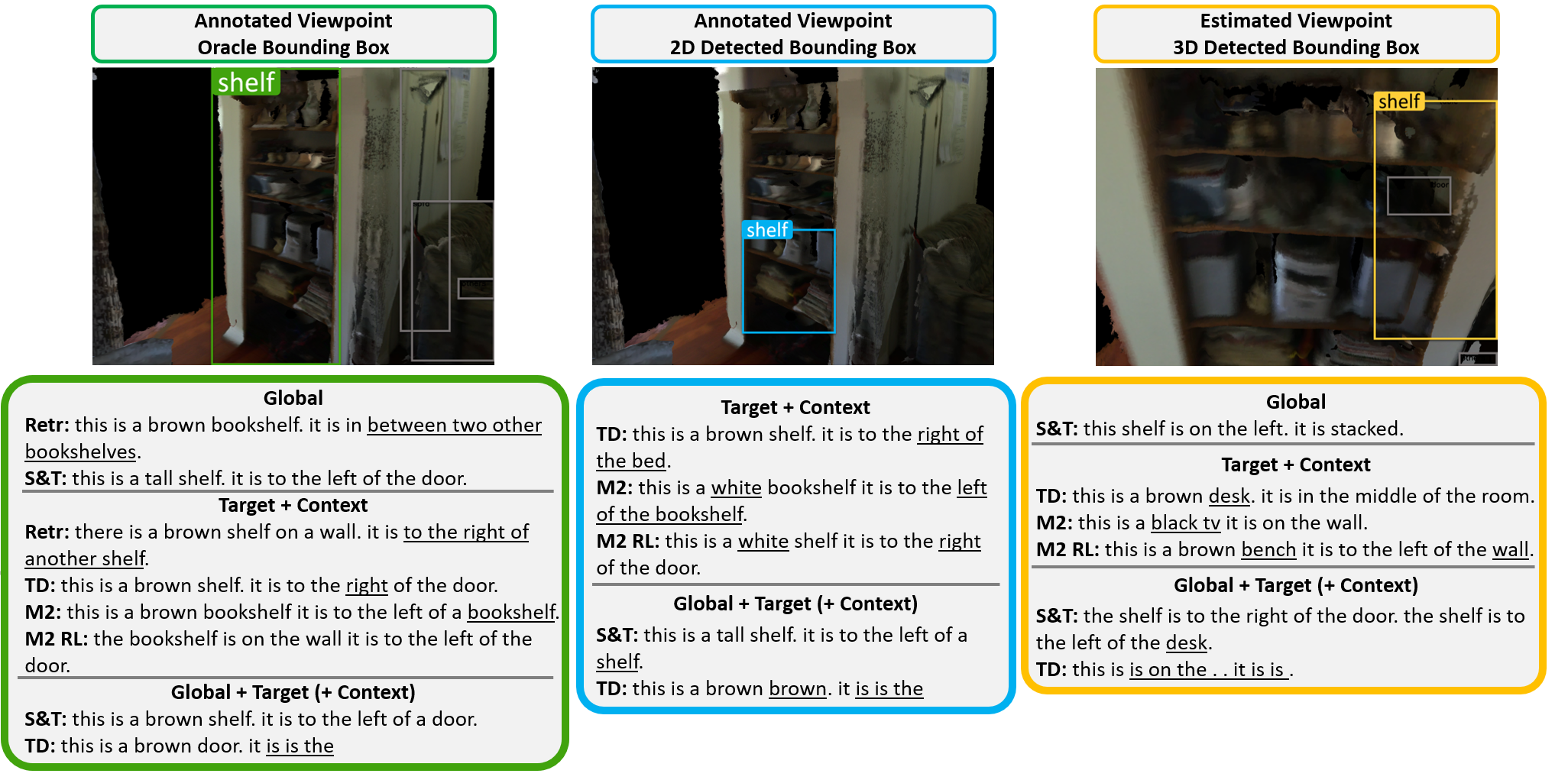}
        \includegraphics[width=\textwidth]{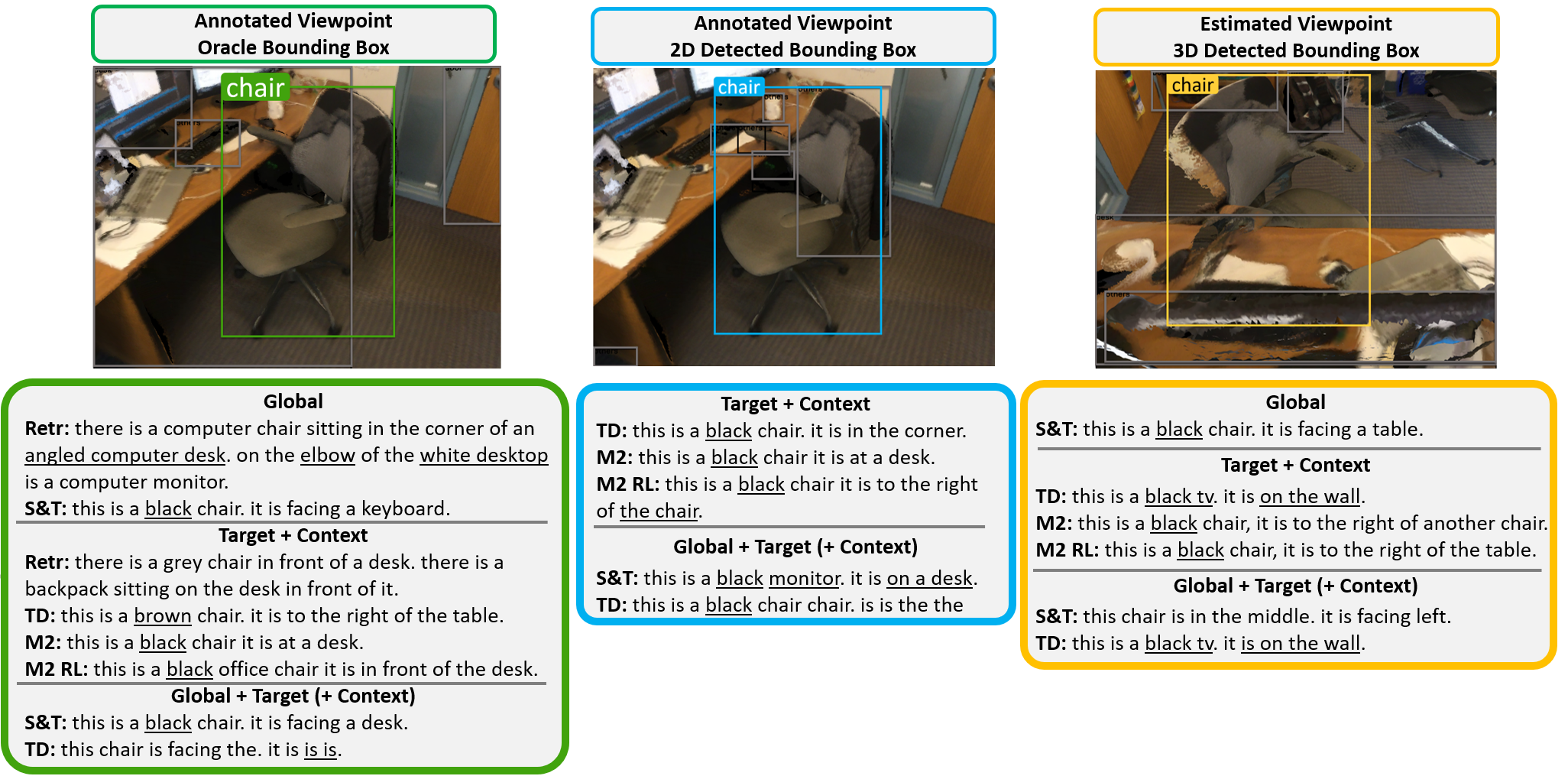}
        \label{fig:q2d_1}
        \caption{Examples of captions generated from 2D rendered frames with oracle bounding boxes (O-left), detected boxes from Mask-RCNN (2DM-middle), and projected bounding boxes from 3D to 2D (3DV-right).  The oracle and Mask-RCNN predictions are from the annotated viewpoint, while the 3D to 2D projection is using an estimated viewpoint.  The bounding box for the target object is shown in color, while the bounding box for the context objects are in gray.  Inaccurate parts of the caption are underscored.}
    %\end{subfigure}
    % \begin{subfigure}{\textwidth}
    %     \centering
    %     \includegraphics[width=\textwidth]{figures/supplements/q2d_2.png}
    %     \label{fig:q2d_2}
    % \end{subfigure}
    % \caption{Examples of captions generated from 2D rendered frames with oracle bounding boxes (O-left), detected boxes from Mask-RCNN (2DM-right), and projected bounding boxes from 3D to 2D (3DV-middle).  The oracle and Mask-RCNN predictions are from the annotated viewpoint, while the 3D to 2D projection is using an estimated viewpoint.  The bounding box for the target object is shown in color, while the bounding box for the context objects are in gray.  Inaccurate parts of the caption are underscored}
    \label{fig:qualitative_2d}
\end{figure*}

\paragraph{Visual features}
For visual features, we experiment with using the global visual features for the entire image (G), features from just the target object (T), and features from the context objects (C).  For object-based features, we rely on object bounding boxes that are either oracle (O), detected using a 2D object detector (2DM), or back-projected from 3D (3DV).  For our 2D detection, We use Mask R-CNN~\citep{he2017mask} with a pre-trained ResNet-101~\citep{he2016deep} as our backbone and then fine-tune it on the ScanRefer training split using rendered viewpoints.

\subsection{Results}
In this section we evaluate our instance segmentation and captioning methods in 2D.

\subsubsection{Object detection and instance segmentation}

We evaluate the model performance on object detection and instance segmentation via \textit{mAP} (mean average precision). Tab.~\ref{tab:detection_and_segmentation} demonstrates our object detection and instance segmentation results.

\subsubsection{Captioning}
We evaluate the captions generated for 2D inputs using the well-established CiDEr~\citep{vedantam2015cider}, BLEU-4~\citep{papineni2002bleu}, METEOR~\citep{banerjee2005meteor} and ROUGE~\citep{lin2004rouge}, abbreviated as C, B-4, M, R, respectively. Tab.~\ref{tab:comp_baseline} shows our captioning experiment results and Fig.~\ref{fig:qualitative_2d} shows examples from the different methods.  Note that the captioning metrics reported here are not comparable to dense captioning metrics reported in the main paper, as these does not take into account the IoU, and we evaluate the predicted caption against the ground truth caption for each respective viewpoint.

Surprisingly, we find that the simple baseline of S\&T outperforms other methods such as the top-down attention (TD) and meshed-memory transformer (M$^2$) on CiDEr and METEOR. We suspect that this is partly due to the limited amount of training data (MSCOCO has 113,287 training images with five captions each while ScanRefer has only 36,665 descriptions in the train split). Thus, for our 2D-based baselines in the main paper, we chose to use S\&T with features from the global image and the target object.

\section{3D to 2D projection details}
\label{sec:3d2dproj_details}

\begin{figure*}[!ht]
    \centering
    \includegraphics[width=\linewidth]{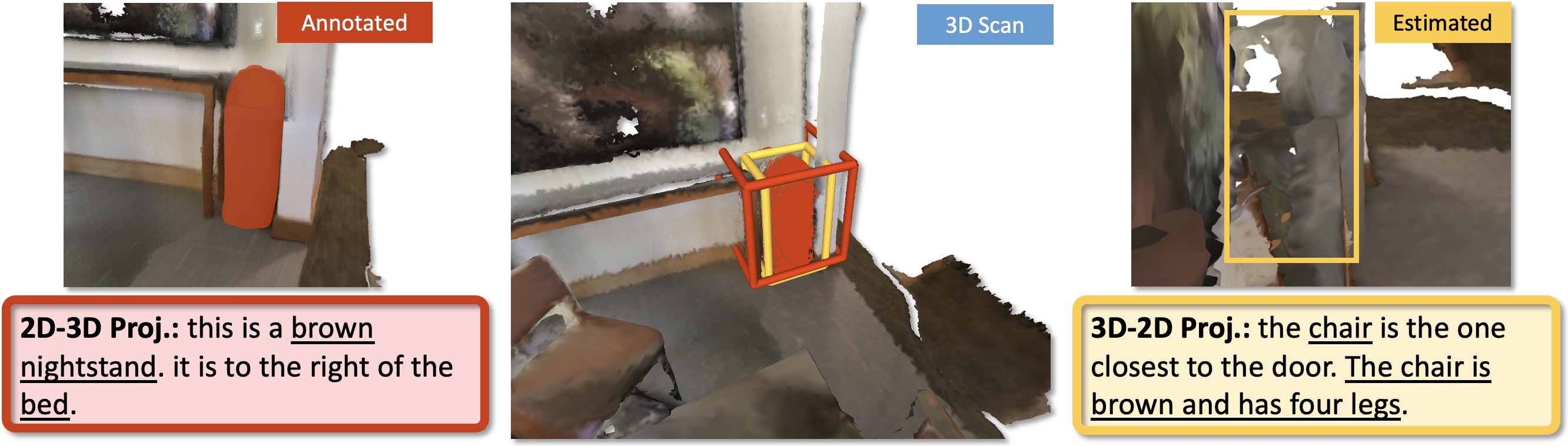}
    \caption{
    Comparison of generated captions based on 2D-3D and 3D-2D projected detections (2D-3D Proj. and 3D-2D Proj respectively). In 2D-3D Proj., we first detect object mask in the rendered annotated viewpoints using Mask R-CNN~\citep{he2017mask} (as shown in the red box on the left), and generate the caption for the detected object. While in 3D-2D Proj., we first detect object bounding boxes in 3D using VoteNet~\citep{qi2019deep}, then estimate a viewpoint for the detected 3D bounding box, and we back-project the detected bounding box to 2D. We then generate the caption based on the estimated viewpoint and the back-projected bounding box (see the yellow box on the right).}
    \label{fig:2D-3D_projection}
\end{figure*}

In order to caption the objects in the images using 3D detected information, we estimate the camera viewpoints from the 3D bounding boxes and  project the 3D bounding boxes to the rendered single-view images for captioning. We show the example in Fig.~\ref{fig:2D-3D_projection}.

\paragraph{Viewpoint estimation from 3D detections.}

We take several heuristics into account to estimate the viewpoints for the detected 3D boxes. To start with, we compute the average distance between the target objects and the recorded viewpoints (1.97 meters). Then, assuming the camera height as 1.70 meters, we compute the horizontal distance between the target objects and the viewpoints (0.99 meter). We randomly pick the points on the circle with the horizontal radius 0.99 meters to the target objects. We repeat the random selection process until the selected viewpoints are inside the scenes and the target objects are visible in the view. 

\paragraph{Projecting 3D detections to the estimated views.}

We derive the camera extrinsics from the estimated viewpoints as we assume the cameras are always targeting at the center of the 3D bounding boxes. We keep the camera intrinsic as in ScanNet. Then, we use these camera parameters to render the single-view images for the 3D scans. The 3D bounding boxes are then projected into the image space as the targets and contexts for generating captions.

\section{Additional 3D captioning results}
\subsection{Additional quantitative analysis}
\label{sec:add_quantitative}

\begin{figure*}[!ht]
    \centering
    \includegraphics[width=\linewidth]{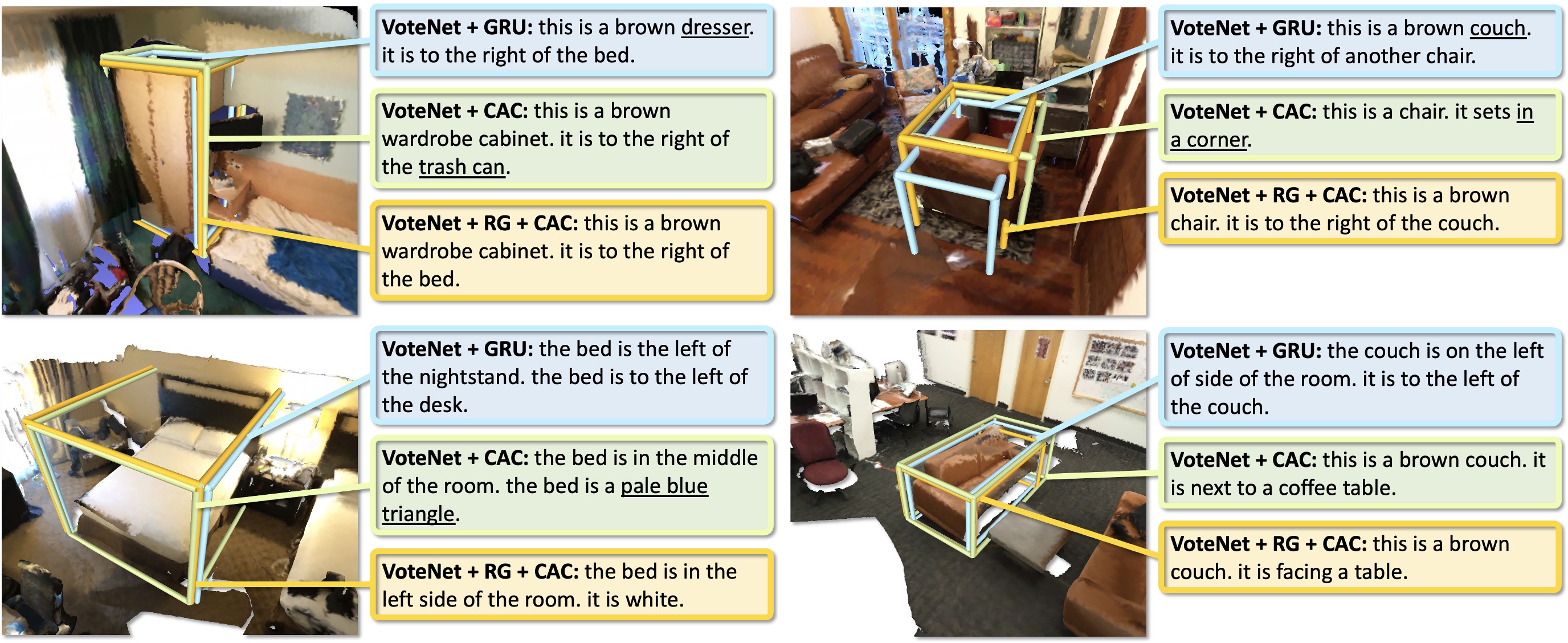}
    \caption{Ablation study with different components in our method: VoteNet~\citep{qi2019deep} + GRU~\citep{chung2014empirical}, which is similar to ``show and tell''~\citet{vinyals2015show}; VoteNet + Context-aware Attention Captioning (CAC); VoteNet + Relational Graph (RG) + Context-aware Attention Captioning (CAC), namely Scan2Cap. We underscore the inaccurate aspects in the descriptions. Image best viewed in color.}
    \label{fig:qualitative_components}
\end{figure*}

\begin{table*}[!ht]
    \centering
    \resizebox{\linewidth}{!}{
        \begin{tabular}{l|cccc|cccc|c}
            \toprule
            & C@0.25IoU & B-4@0.25IoU & M@0.25IoU & R@0.25IoU & C@0.5IoU & B-4@0.5IoU & M@0.5IoU & R@0.5IoU & mAP@0.5IoU \\
            \midrule
            Ours (xyz) & 47.21 & 29.41 & 24.89 & 50.74 & 32.94 & 20.63 & 21.10 & 41.58 & 27.45 \\
            Ours (xyz+rgb) & 49.36 & 32.88 & 25.52 & 54.20 & 33.41 & 21.61 & \textbf{22.12} & 43.61 & 27.52 \\
            Ours (xyz+rgb+normal) & 53.73 & \textbf{34.25} & 26.14 & 54.95 & 35.20 & 22.36 & 21.44 & 43.57 & 29.13 \\
            Ours (xyz+multiview) & 54.94 & 32.73 & 25.90 & 53.51 & 36.89 & 21.77 & 21.39 & 42.83 & 31.43 \\ 
            Ours (xyz+multiview+normal)& \textbf{56.82} & 34.18 & \textbf{26.29} & \textbf{55.27} & \textbf{39.08} & \textbf{23.32}  & 21.97& \textbf{44.78} & \textbf{32.21} \\
            \bottomrule
        \end{tabular}
    }
    \caption{Ablation study with different features. We compute standard captioning metrics with respect to the percentage of the predicted bounding box whose IoU with the ground truth are greater than 0.25 and 0.5. The higher the better.}
    \label{tab:comp_feature}
    \vspace{1cm}
\end{table*}

\paragraph{Do other 3D features help?}
We include colors and normals from the ScanNet meshes to the input point cloud features and compare performance against networks trained without them. As displayed in Tab.~\ref{tab:comp_feature}, our architecture trained with geometry, multi-view features and normal vectors (xyz+multiview+normal) achieves the best performance among all ablations.  This matches the feature ablation from ScanRefer~\cite{chen2020scanrefer}.

\subsection{Additional qualitative analysis}
\label{sec:add_qualitative}

\paragraph{Do graph and attention help with captioning?}
We compare our model (VoteNet+RG+CAC) with the basic description generation component (VoteNet+GRU) introduced in~\citet{vinyals2015show} and our model equipped only with the context-aware attention captioning (VoteNet+CAC). As shown in Fig.~\ref{fig:qualitative_components}, though all three methods produce good bounding boxes (IoU$>$0.5), VoteNet+GRU makes mistakes when describing the target objects. VoteNet+CAC refers to the target and the objects nearby in the scene, but still fails to correctly reveal the relative spatial relationships. In contrast, VoteNet+RG+CAC can properly handle the interplay of describing the target appearance and the relative spatial relationships in the local environment.

\end{document}